# Orthographic Syllable as basic unit for SMT between Related Languages

**Anoop Kunchukuttan, Pushpak Bhattacharyya**
Center For Indian Language Technology,
Department of Computer Science & Engineering
Indian Institute of Technology Bombay
`{anoopk,pb}@cse.iitb.ac.in`

## Abstract

We explore the use of the *orthographic syllable*, a variable-length consonant-vowel sequence, as a basic unit of translation between *related* languages which use abugida or alphabetic scripts. We show that orthographic syllable level translation significantly outperforms models trained over other basic units (word, morpheme and character) when training over small parallel corpora.

## 1 Introduction

*Related languages* exhibit lexical and structural similarities on account of sharing a **common ancestry** (Indo-Aryan, Slavic languages) or being in **prolonged contact** for a long period of time (Indian subcontinent, Standard Average European linguistic areas) (Bhattacharyya et al., 2016). Translation between *related* languages is an important requirement due to substantial government, business and social communication among people speaking these languages. However, most of these languages have few parallel corpora resources, an important requirement for building good quality SMT systems.

Modelling the lexical similarity among related languages is the key to building good-quality SMT systems with limited parallel corpora. *Lexical similarity* implies that the languages share many words with the similar form (spelling/pronunciation) and meaning *e.g.* `blindness` is `andhapana` in Hindi, `aandhaLepaNaa` in Marathi. These words could be cognates, lateral borrowings or loan words from other languages. Translation for such words can be achieved by sub-word level transformations. For instance, lexical similarity can be modelled in the standard SMT pipeline by transliteration of words while decoding (Durrani et al., 2010) or post-processing (Nakov and Tiedemann, 2012; Kunchukuttan et al., 2014).

A different paradigm is to drop the notion of word boundary and consider the character n-gram as the basic unit of translation (Vilar et al., 2007; Tiedemann, 2009a). Such character-level SMT bas been explored for closely related languages like *Bulgarian-Macedonian, Indonesian-Malay* with modest success, with the short context of unigrams being a limiting factor (Tiedemann, 2012). The use of character n-gram units to address this limitation leads to data sparsity for higher order n-grams and provides little benefit (Tiedemann and Nakov, 2013).

In this work, we present a linguistically motivated, variable length unit of translation — **orthographic syllable (OS)** — which provides more context for translation while limiting the number of basic units. The OS consists of one or more consonants followed by a vowel and is inspired from the *akshara*, a consonant-vowel unit, which is the fundamental organizing principle of Indic scripts (Sproat, 2003; Singh, 2006). It can be thought of as an *approximate syllable* with the onset and nucleus, but no coda. While true syllabification is hard, orthographic syllabification can be easily done. Atreya et al. (2016) and Ekbal et al. (2006) have shown that the OS is a useful unit for transliteration involving Indian languages.

We show that orthographic syllable-level trans-lation significantly outperforms character-level and strong word-level and morpheme-level baselines over multiple related language pairs (Indian as well as others). Character-level approaches have been previously shown to work well for language pairs with high lexical similarity. Our major finding is that OS-level translation outperforms other approaches even when the language pairs have relatively less lexical similarity or belong to different language families (but have sufficient contact relation).

## 2 Orthographic Syllabification

The *orthographic syllable* is a sequence of one or more consonants followed by a vowel, *i.e* a C$^+$V unit. We describe briefly procedures for orthographic syllabification of Indian scripts and non-Indic alphabetic scripts. Orthographic syllabification cannot be done for languages using *logographic* and *abjad* scripts as these scripts do not have vowels.

**Indic Scripts:** Indic scripts are *abugida* scripts, consisting of consonant-vowel sequences, with a consonant core ($C^+$) and a dependent vowel (*matra*). If no vowel follows a consonant, an implicit *schwa* vowel [IPA: ə] is assumed. Suppression of *schwa* is indicated by the *halanta* character following a consonant. This script design makes for a straightforward syllabification process as shown in the following example. *e.g.* लक्ष्मी ($\frac{lakShamI}{CVCCVCV}$) is segmented as ल क्ष मी ($\frac{la\ kSha\ mI}{CV\ CCV\ CV}$). There are two exceptions to this scheme: (i) Indic scripts distinguish between dependent vowels (vowel diacritics) and independent vowels, and the latter will constitute an OS on its own. *e.g.* मुम्बई ($mumbaI$) → मु म्ब ई ($mu\ mba\ I$) (ii) The characters *anusvaara* and *chandrabindu* are part of the OS to the left if they represents nasalization of the vowel/consonant or start a new OS if they represent a nasal consonant. Their exact role is determined by the character following the *anusvaara*. See Appendix A for details.

**Non-Indic Alphabetic Scripts:** We use a simpler method for the alphabetic scripts used in our experiments (Latin and Cyrillic). The OS is identified by a C$^+$V$^+$ sequence. *e.g. lakshami*→*la ksha mi*, *mumbai*→*mu mbai*. The OS could contains multiple terminal vowel characters representing long vowels (*oo* in *cool*) or diphthongs (*ai* in *mumbai*). A vowel starting a word is considered to be an OS.

| Basic Unit | Example | Transliteration |
|---|---|---|
| Word | घरासमोरचा | gharAsamoracA |
| Morph Segment | घरा समोर चा | gharA samora cA |
| Orthographic Syllable | घ रा स मो र चा | gha rA sa mo r acA |
| Character unigram | घ र ा स म ो र च ा | gha r A sa m o r a c A |
| Character 3-gram | घरा समो रचा | gharA samo rcA |

*something that is in front of home:* ghara=home, samora=front, cA=of

Table 1: Various translation units for a Marathi word

## 3 Translation Models

We compared the orthographic syllable level model (O) with models based on other translation units that have been reported in previous work: word (W), morpheme (M), unigram (C) and trigram characters. Table 1 shows examples of these representations.

The first step to build these translation systems is to transform sentences to the correct representation. Each word is segmented as the per the unit of representation, punctuations are retained and a special *word boundary marker* character (_) is introduced to indicate word boundaries as shown here:

W:  राजू , घराबाहेर जाऊ नको .
O:  रा जू _ , _ घ रा बा हे र _ जा ऊ _ न को _ .

For all units of representation, we trained phrase-based SMT (PBSMT) systems. Since related languages have similar word order, we used distance based distortion model and monotonic decoding. For character and orthographic syllable level models, we use higher order (10-gram) languages models since data sparsity is a lesser concern due to small vocabulary size (Vilar et al., 2007). As suggested by Nakov and Tiedemann (2012), we used word-level tuning for character and orthographic syllable level models by post-processing n-best lists in each tuning step to calculate the usual word-based BLEU score.

While decoding, the word and morpheme level systems will not be able to translate OOV words. Since the languages involved share vocabulary, we transliterate the untranslated words resulting in the post-edited systems W$_X$ and M$_X$ corresponding to the systems W and M respectively. Following decoding, we used a simple method to regenerate words from sub-word level units: Since we represent word boundaries using a word boundary marker, we

| IA→IA | | DR→DR | | IA→DR | |
|---|---|---|---|---|---|
| ben-hin | 52.30 | mal-tam | 39.04 | hin-mal | 33.24 |
| pan-hin | 67.99 | tel-mal | 39.18 | **DR→IA** | |
| kok-mar | 54.51 | | | mal-hin | 33.24 |

*IA: Indo-Aryan, DR: Dravidian*

Table 2: Language pairs used in experiments along with Lexical Similarity between them, in terms of LCSR between training corpus sentences

simply concat the output units between consecutive occurrences of the marker character.

## 4 Experimental Setup

**Languages:** Our experiments primarily concentrated on multiple language pairs from the two major language families of the Indian sub-continent (Indo-Aryan branch of Indo-European and Dravidian). These languages have been in contact for a long time, hence there are many lexical and grammatical similarities among them, leading to the sub-continent being considered a *linguistic area* (Emeneau, 1956). Specifically, there is overlap between the vocabulary of these languages to varying degrees due to cognates, language contact and loanwords from Sanskrit (throughout history) and English (in recent times). Table 2 lists the languages involved in the experiments and provides an indication of the lexical similarity between them in terms of the Longest Common Subsequence Ratio (LCSR) (Melamed, 1995) between the parallel training sentences at character level. All these language have a rich inflectional morphology with Dravidian languages, and Marathi and Konkani to some degree, being agglutinative. *kok-mar* and *pan-hin* have a high degree of lexical similarity.

**Dataset:** We used the multilingual ILCI corpus for our experiments (Jha, 2012), consisting of a modest number of sentences from tourism and health domains. The data split is as follows – *training: 44,777, tuning 1K, test: 2K* sentences. Language models for word-level systems were trained on the target side of training corpora plus monolingual corpora from various sources [hin: 10M (Bojar et al., 2014), tam: 1M (Ramasamy et al., 2012), mar: 1.8M (news websites), mal: 200K (Quasthoff et al., 2006) sentences]. We used the target language side of the parallel corpora for character, morpheme and OS level LMs.

**System details:** PBSMT systems were trained using the *Moses* system (Koehn et al., 2007), with the *grow-diag-final-and* heuristic for extracting phrases, and Batch MIRA (Cherry and Foster, 2012) for tuning (default parameters). We trained 5-gram LMs with Kneser-Ney smoothing for word and morpheme level models and 10-gram LMs for character and OS level models. We used the *BrahmiNet* transliteration system (Kunchukuttan et al., 2015) for post-editing, which is based on the transliteration Module in Moses (Durrani et al., 2014). We used unsupervised morphological segmenters trained with *Morfessor* (Virpioja et al., 2013) for obtaining morpheme representations. The unsupervised morphological segmenters were trained on the ILCI corpus and the Leipzig corpus (Quasthoff et al., 2006).The morph-segmenters and our implementation of orthographic syllabification are made available as part of the *Indic NLP Library*[1].

**Evaluation:** We use BLEU (Papineni et al., 2002) and Le-BLEU (Virpioja and Grönroos, 2015) for evaluation. Le-BLEU does fuzzy matches of words and hence is suitable for evaluating SMT systems that perform transformation at the sub-word level.

## 5 Results and Discussion

This section discusses the results on Indian and non-Indian languages and cross-domain translation.

**Comparison of Translation Units:** Table 3 compares the BLEU scores for various translation systems. The orthographic syllable level system is clearly better than all other systems. It significantly outperforms the character-level system (by 46% on an average). The character-based system is competitive only for highly lexically similar language pairs like *pan-hin* and *kok-mar*. The system also outperforms two strong baselines which address data sparsity: (a) a word-level system with transliteration of OOV words (10% improvement), (b) a morph-level system with transliteration of OOV words (5% improvement). The OS-level representation is more beneficial when morphologically rich languages are involved in translation. Significantly, OS-level translation is also the best system for translation between languages of different language families. The Le-BLEU scores also show the same trend as BLEU scores (See Appendix B). There are a very small number of untranslated OSes, which we handled by simple mapping of untranslated characters from source to target script. This barely increased translation accuracy (0.02% increase in BLEU score).

---
[1] `http://anoopkunchukuttan.github.io/indic_nlp_library`

|         | W     | $W_X$ | M     | $M_X$ | C     | O     |
|---------|-------|-------|-------|-------|-------|-------|
| ben-hin | 31.23 | 32.79 | 32.17 | 32.32 | 27.95 | **33.46** |
| pan-hin | 68.96 | 71.71 | 71.29 | 71.42 | 71.26 | **72.51** |
| kok-mar | 21.39 | 21.90 | 22.81 | 22.82 | 19.83 | **23.53** |
| mal-tam | 6.52  | 7.01  | 7.61  | 7.65  | 4.50  | **7.86** |
| tel-mal | 6.62  | 6.94  | 7.86  | 7.89  | 6.00  | **8.51** |
| hin-mal | 8.49  | 8.77  | 9.23  | 9.26  | 6.28  | **10.45** |
| mal-hin | 15.23 | 16.26 | 17.08 | 17.30 | 12.33 | **18.50** |

Table 3: Results - ILCI corpus (% BLEU). The reported scores are:- **W**: word-level, $W_X$: word-level followed by transliteration of OOV words, **M**: morph-level, $M_X$: morph-level followed by transliteration of OOV morphemes, **C**: character-level, **O:** orthographic syllable. The values marked in bold indicate the best scores for the language pair.

|         | C    | O    | M    | W    |
|---------|------|------|------|------|
| ben-hin | 0.71 | 0.63 | 0.58 | 0.40 |
| pan-hin | 0.72 | 0.70 | 0.64 | 0.50 |
| kok-mar | 0.74 | 0.68 | 0.63 | 0.64 |
| mal-tam | 0.77 | 0.71 | 0.56 | 0.46 |
| tel-mal | 0.78 | 0.65 | 0.52 | 0.45 |
| hin-mal | 0.79 | 0.59 | 0.46 | -0.02 |
| mal-hin | 0.71 | 0.61 | 0.45 | 0.37 |

Table 4: Pearson's correlation coefficient between lexical similarity and translation accuracy (both in terms of LCSR at character level). *This was computed over the test set between: (ii) sentence level lexical similarity between source and target sentences and (ii) sentence level translation match between hypothesis and reference.*

|         | $W_X$ | $M_X$ | C | O |
|---------|-------|-------|---|---|
| ben-hin | *Corpus not available* | | | |
| pan-hin | 61.56 | **59.75** | 58.07 | 58.48 |
| kok-mar | 19.32 | 18.32 | 17.97 | **19.65** |
| mal-tam | **5.88** | 6.02 | 4.12 | **5.88** |
| tel-mal | 3.19 | **4.07** | 3.11 | 3.77 |
| hin-mal | 5.20 | 6.00 | 3.85 | **6.26** |
| mal-hin | 9.68 | 11.44 | 8.42 | **13.32** |

Table 5: Results: Agricuture Domain (% BLEU)

**Why is OS better than other units?:** The improved performance of OS level representation can be attributed to the following factors:

One, the number of basic translation units is limited and small compared to word-level and morpheme-level representations. For word-level representation, the number of translation units can increase with corpus size, especially for morphologically rich languages which leads to many OOVs. Thus, OS-level units **address data sparsity**.

Two, while character level representation too does not suffer from data sparsity, we observe that the translation accuracy is highly correlated to lexical similarity (Table 4). The high correlation of character-level system and lexical similarity explains why character-level translation performs nearly as well other methods for language pairs which have high lexical similarity, but performs badly otherwise. On the other hand, the OS-level representation has lesser correlation with lexical similarity and sits somewhere between character-level and word/morpheme level systems. Hence it is able to make **generalizations beyond simple character level mappings**. We observed that OS-level representation was able to correctly generate words whose translations are not cognate with the source language. This is an important property since function words and suffixes tend to be less similar lexically across languages.

Can improved translation performance be explained by longer basic translation units? To verify this, we trained translation systems with character trigrams as basic units. We chose trigrams since the average length of the OS was 3-5 characters for the languages we tested with. The translation accuracies were far less than even unigram representation. The number of unique basic units was about 8-10 times larger than orthographic syllables, thus making data sparsity an issue again. So, improved translation performance **cannot be attributed to longer n-gram units alone**.

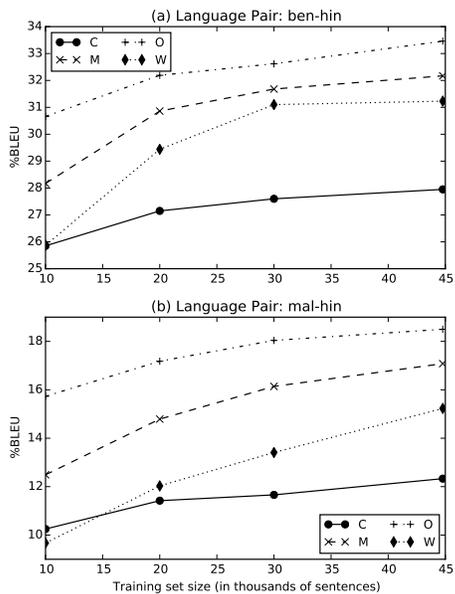

Figure 1: Effect of training data size on translation accuracy for different basic units

|  | Corpus Stats | Lex-Sim | W | C | O |
|---|---|---|---|---|---|
| bul-mac | (150k,1k,2k) | 62.85 | 21.20 | 20.61 | **21.38** |
| dan-swe | (150k,1k,2k) | 63.39 | 35.13 | 35.36 | **35.46** |
| may-ind | (137k,1k,2k) | 73.54 | 61.33 | 60.50 | **61.24** |

Table 6: Translation among non-Indic languages (%BLEU). Corpus Stats show (train,tune,test) split

**Robustness to Domain Change:** We also tested the translation models trained on tourism & health domains on an agriculture domain test set of 1000 sentences. In this cross-domain translation scenario too, the OS level model outperforms most units of representation. The only exceptions are the *pan-hin* and *tel-mal* language pairs for the system $\mathbf{M}_X$ (accuracies of the OS-level system are within 10% of the $\mathbf{M}_X$ system). Since the word leve model depends on coverage of the lexicon, it is highly domain dependent, whereas the sub-word units are not. So, even unigram-level models outperform word-level models in a cross-domain setting.

**Experiments with non-Indian languages:** Table 6 shows the corpus statistics and our results for translation between some related non-Indic language pairs (Bulgarian-Macedonian, Danish-Swedish, Malay-Indonesian). OS level representation outperforms character and word level representation, though the gains are not as significant as Indic language pairs. This could be due to short length of sentences in training corpus [OPUS movie subtitles (Tiedemann, 2009b)] and high lexical similarity between the language pairs. Further experiments between less lexically related languages on general parallel corpora will be useful.

**Effect of training data size:** For different training set sizes, we trained SMT systems with various representation units (Figure 1 shows the learning curves for two language pairs). OS level models are consistently better than character level models even for small training corpora. While the OS-level models benefit from larger training corpora, the character-level models show only small benefits. For smaller corpora, the OS level model shows even larger performance gains over the corresponding word and morph level models. The performance gap between word/morph-level and OS-level shows a decrease with increase in training data.

## 6 Conclusion & Future Work

We focus on the task of translation between *related languages*. This aspect of MT research is important to make available translation technologies to language pairs with limited parallel corpus, but huge potential translation requirements. We propose the use of the *orthographic syllable*, a variable-length, linguistically motivated, approximate syllable, as a basic unit for translation between related languages. We show that it significantly outperforms other units of representation, over multiple language pairs, spanning different language families, with varying degrees of lexical similarity and is robust to domain changes too. This opens up the possibility of further exploration of sub-word level translation units *e.g.* segments learnt using byte pair encoding (Sennrich et al., 2016).

## Acknowledgments

We thank Arjun Atreya for inputs regarding orthographic syllables. We thank the Technology Development for Indian Languages (TDIL) Programme and the Department of Electronics & Information Technology, Govt. of India for their support.

## Algorithm 1 Orthographic Syllabification of Indic Scripts

```
1: function ORTH-SYLLABIFY(W)
     W: word to be syllabified, SW: syllabified word (spaces added at boundaries)
2:     SW ← ''                                                          ▷ Initialized to empty string
3:     for i ← 1, len(W) do                                              ▷ Iterate over characters
4:         SW ← SW + W[i]
5:         if is_vowel(W[i]) and not is_nasalizer(W[i+1], W[i+2]) then
6:             SW ← SW + ' '                                             ▷ Add space
7:         else if is_consonant(W[i]) then
8:             if is_depvowel(W[i+1]) or is_consonant(W[i+1]) then
9:                 SW ← SW + ' '
10:            else if not is_nasalizer(W[i+1], W[i+2]) then
11:                SW ← SW + ' '
12:            end if
13:        end if
14:    end for
15:    return SW
16: end function

17: function is_nasalizer(c_1, c_2)
      c_1: character to check, c_2: character following c_1
18:    is_nasal ← is_anusvar(c_1) or is_chandrabindu(c_1)                 ▷ Checks for certain characters
19:    return is_nasal and not is_plosive(c_2)
20: end function
```

## A. Pseudo-code for Orthographic Syllabification of Indic Scripts

A minimal pseudo-code for orthographic syllabification of Indic scripts is shown in Algorithm 1.

## B. Le-BLEU Scores

In this appendix, we report and briefly analyse the Le-BLEU scores for all the experiments. Tables 7 shows the Le-BLEU scores for Indian language pairs. Table 8 shows the Le-BLEU scores for cross-domain translation. We see that the Le-BLEU score also clearly shows that the OS level representation is better than other representations. Especially in the cross-domain scenario, compared to the BLEU scores the Le-BLEU scores clearly indicate that superiority of the OS level representation.

|  | W | $W_X$ | M | $M_X$ | C | O |
|---|---|---|---|---|---|---|
| ben-hin | 67.63 | 71.03 | 70.77 | 71.18 | 67.22 | **71.66** |
| pan-hin | 86.81 | 90.13 | 89.95 | 90.34 | 90.45 | **90.66** |
| kok-mar | 63.97 | 63.74 | 65.95 | 65.85 | 63.22 | **67.05** |
| mal-tam | 31.87 | 40.72 | 40.92 | 43.43 | 31.11 | **44.19** |
| tel-mal | 31.35 | 37.92 | 38.33 | 39.34 | 34.62 | **43.79** |
| hin-mal | 40.48 | 43.89 | 43.55 | 44.05 | 32.41 | **46.65** |
| mal-hin | 46.03 | 52.28 | 52.8 | 54.17 | 44.38 | **55.5** |

Table 7: Results: ILCI Corpus (% Le-BLEU)

|  | $W_X$ | $M_X$ | C | O |
|---|---|---|---|---|
| ben-hin | *Corpus not available* | | | |
| pan-hin | 87.44 | **86.95** | 86.57 | 86.73 |
| kok-mar | 63.9 | 63.82 | 64.75 | **66.68** |
| mal-tam | 41.27 | 42.75 | 30.09 | **44.62** |
| tel-mal | 26.31 | 24.88 | 24.61 | **29.26** |
| hin-mal | 38.11 | 36.2 | 28.13 | **39.3** |
| mal-hin | 48.98 | 51.25 | 43.91 | **54.57** |

Table 8: Results: Agricuture Domain (% Le-BLEU)